# Computing upper and lower bounds on likelihoods in intractable networks

**Tommi S. Jaakkola and Michael I. Jordan**
Department of Brain and Cognitive Sciences
Massachusetts Institute of Technology
Cambridge, MA 02139
*{tommi,jordan}@psyche.mit.edu*

## Abstract

We present deterministic techniques for computing upper and lower bounds on marginal probabilities in sigmoid and noisy-OR networks. These techniques become useful when the size of the network (or clique size) precludes exact computations. We illustrate the tightness of the bounds by numerical experiments.

## 1 INTRODUCTION

A graphical model provides an explicit representation of qualitative dependencies among the variables associated with the nodes of the graph (Pearl, 1988). Numerical specification of these dependencies in the form of potentials or probability tables enables quantitative computation of beliefs about the values of the variables on the basis of acquired evidence. The computations involved, i.e., propagation of beliefs, can be handled by now standard exact methods (Lauritzen & Spiegelhalter, 1988, Jensen et al. 1990). Junction trees serve as representational platforms for these exact probabilistic calculations and are constructed from directed graphical representations via moralization and triangulation. Although powerful in utilizing the structure of the underlying networks, junction trees may, in some cases, contain cliques that are prohibitively large. In such cases it is desirable to develop approximate methods that bound the marginal probabilities. As an alternative to Monte Carlo methods, which approximate marginal probabilities in a stochastic sense, we develop deterministic methods that yield strict lower and upper bounds for the marginals. These bounds together yield interval bounds on the desired probabilities. Although the problem of finding such intervals to predescribed accuracy is NP-hard (Dagum and Luby, 1993), bounds that can be computed efficiently may nevertheless yield intervals that are accurate enough to be useful in practice.

Large clique sizes (arising from dense connectivity) lead not only to long execution times but also involve exponentially many parameters that must be assessed or learned. The latter issue is generally addressed via parsimonious representations such as the logistic sigmoid (Neal, 1992) or the noisy-OR function (Pearl, 1988). We consider both of these representations in the current paper. We stay within a directed framework and thereby retain the compactness of these representations throughout our inference and estimation algorithms.

Saul et al. (1996) derived a rigorous *lower* bound for sigmoid belief networks. We complete the picture here by developing the missing *upper* bounds for sigmoid networks. We also develop both upper and lower bounds for noisy-OR networks. While the lower bounds we obtain are applicable to generic network structures, the upper bounds are currently restricted to two-level networks. Although a serious restriction, there are nonetheless many potential applications for such upper bounds, including the probabilistic reformulation of the QMR knowledge base (Shwe et al., 1991). We emphasize finally that our focus in this paper is on techniques of bounding rather than on all-encompassing inference algorithms; tailoring the bounds for specific problems or merging them with exact methods may yield a considerable advantage.

The paper is structured as follows. Section 2 introduces sigmoid belief networks, develops the techniques for upper and lower bounds, and gives preliminary numerical analysis of the accuracy of the bounds. Section 3 is devoted to the analogous results for noisy-OR networks. In section 4 we summarize the results and describe some future work.

## 2 SIGMOID BELIEF NETWORKS

Sigmoid belief networks are (directed) probabilistic networks defined over binary variables $S_1, \ldots, S_n$. The joint distribution for the variables has the usual decompositional structure:

$$P(S_1, \ldots, S_n|\theta) = \prod_i P(S_i|\text{pa}[i], \theta) \qquad (1)$$

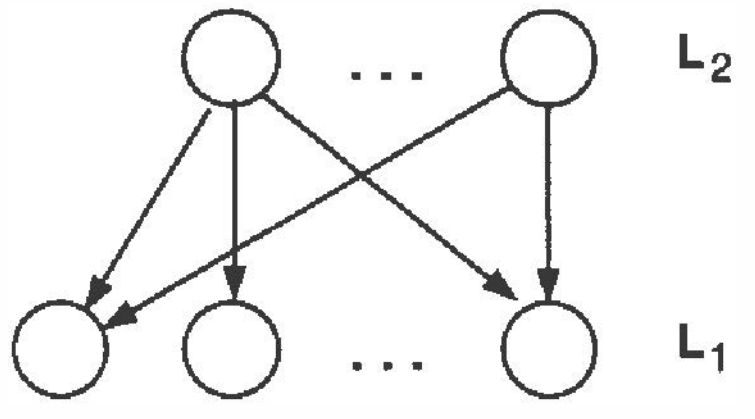

Figure 1: Two level (bipartite) network.

The conditional probabilities, however, take a particular form given by

$$
\begin{aligned}
P(S_i|\mathrm{pa}[i], \theta) &= \\
&= g(\textstyle\sum_{j\in \mathrm{pa}[i]} \theta_{ij} S_j)^{S_i} [1 - g(\sum_{j\in \mathrm{pa}[i]} \theta_{ij} S_j)]^{1-S_i} \quad (2) \\
&= g((2S_i - 1)\textstyle\sum_{j\in \mathrm{pa}[i]} \theta_{ij} S_j) \quad (3)
\end{aligned}
$$

where $g(x) = 1/(1+\exp(-x))$ is the logistic function (also called a "sigmoid" function based on its graphical shape; see Figure 6). The parameters specifying these conditional probabilities are the real valued "weights" $\theta_{ij}$. We note that the choice of this dependency model is not arbitrary but is rooted in logistic regression in statistics (McCullagh & Nelder, 1983). Furthermore, this form of dependency corresponds to the assumption that the odds from each parent of a node combine multiplicatively; the weights $\theta_{ij}$ in this interpretation bear a relation to log-odds.

In the remainder of this section we present techniques for computing upper and lower bounds on marginal probabilities in sigmoid networks. We note that any successful instantiation of evidence in these networks relies on the ability to estimate such marginals. The upper bounds that we derive are restricted to two-level (bipartite) networks while the lower bounds are valid for arbitrary network structures.

## 2.1 UPPER BOUND FOR SIGMOID NETWORK

We restrict our attention to two-level directed architectures. The joint probability for this class of models can be written as

$$
\begin{aligned}
P(S_1,\ldots,S_n|\theta) &= \prod_{i\in L_1} g((2S_i-1)\textstyle\sum_{j\in \mathrm{pa}[i]} \theta_{ij} S_j) \\
&\quad \times \prod_{j\in L_2} P(S_j|\theta_j) \quad (4)
\end{aligned}
$$

where $L_1$ and $L_2$ signify the two layers of a bipartite graph with connections from $L_2$ to $L_1$ (see Figure 1).

To compute the marginal probability of a set of variables in these networks we note that (i) any variables in layer $L_2$ included in this set only reduce the complexity of the calculations, and (ii) the form of the architecture makes those variables in $L_1$ that are excluded from the desired marginal set inconsequential. We will thus adopt a simplifying notation in which the marginal set consists of all and only the variables in $L_1$. Thus, the goal is to compute

$$
P(\{S_i\}_{i\in L_1}|\theta) = \sum_{\{S_j\}_{j\in L_2}} P(S_1,\ldots,S_n|\theta) \quad (5)
$$

Given our assumption that computing the marginal probability is intractable, we seek an upper bound instead. Let us briefly outline our strategy. The goal is to simplify the joint distribution such that the marginalization across $L_2$ can be accomplished efficiently, while maintaining at all times a rigorous upper bound on the desired marginal probability. Our approach is to introduce additional parameters into the problem (known as "variational parameters") that can factorize the joint distribution over the variables in $L_2$. Thus we first find a "variational" form for the joint distribution. As we will see below this type of variational form can be obtained by combining variational representations for each sigmoid function in our probability model. Although the variational forms are exact they can be turned into upper bounds by not carrying out the minimizations involved and instead fixing the variational parameters. It is precisely this fixing that leads to the factorization of the joint distribution and consequently allows the marginalization to be carried out efficiently. We note finally that the variational parameters that are kept fixed during the marginalization can be employed afterwards to optimize the bound. In essence, this amounts to exchanging the order of the marginalization and the variational minimization.

To derive the upper bound we first make use of the following variational transformation of the sigmoid function (see appendix A):

$$
g(x) = \frac{1}{1+e^{-x}} = \min_{\xi\in[0,1]} e^{\xi x - H(\xi)} \quad (6)
$$

where $H(\cdot)$ is the binary entropy function. Inserting this transformation into the probability model we find

$$
\begin{aligned}
P(S_1,\ldots,S_n|\theta) &= \\
&= \prod_{i\in L_1} \min_{\xi_i}\left\{ e^{-H(\xi_i)+\xi_i(2S_i-1)\sum_j \theta_{ij}S_j} \right\} \prod_{j\in L_2} P(S_j|\theta_j) \\
&= \min_{\xi}\left\{ e^{-\sum_{i\in L_1} H(\xi_i)} \times \right. \\
&\quad \left. \times \prod_{j\in L2} \left[ e^{\sum_{i\in L_1} \xi_i(2S_i-1)\theta_{ij}} \right]^{S_j} P(S_j|\theta_j) \right\} \quad (7) \\
&\stackrel{def}{=} \min_{\xi}\{\, \tilde{P}(S_1,\ldots,S_n|\theta,\xi)\,\} \quad (8)
\end{aligned}
$$

where we have pulled the minimizations outside and combined the terms that depend on each of the variables $S_j$ in $L_2$. This reorganization shows that $\tilde{P}(S_1,\ldots,S_n|\theta,\xi)$ (defined implicitly) factorizes across $\{S_j\}_{j\in L_2}$ (i.e. across the variables that we need to marginalize over). Thus for any fixed values of the variational parameters, the marginalization can be performed efficiently. We may therefore obtain a simple closed form upper bound on the marginal probability by exchanging the order of the summation and the variational minimization:

$$P(\{S_i\}_{i\in L_1}|\theta) = \sum_{\{S_j\}_{j\in L_2}} P(S_1,\ldots,S_n|\theta) \quad (9)$$

$$= \sum_{\{S_j\}_{j\in L_2}} \min_{\xi}\{\ \tilde{P}(S_1,\ldots,S_n|\theta,\xi)\ \} \quad (10)$$

$$\leq \min_{\xi} \sum_{\{S_j\}_{j\in L_2}} \tilde{P}(S_1,\ldots,S_n|\theta,\xi) \quad (11)$$

$$= \min_{\xi}\left\{ e^{-\sum_{i\in L_1} H(\xi_i)} \times \prod_{j\in L_2}\left(P(S_j=1|\theta_j)e^{\sum_{i\in L_1}\xi_i(2S_i-1)\theta_{ij}} + P(S_j=0|\theta_j)\right)\right\} \quad (12)$$

We state here a few facts about the bound (mostly without proof): (i) The bound can never be greater than one since one is always achieved by setting all $\xi$ to zero, (ii) the bound becomes exact in the limit of small parameter values, and (iii) for fixed prior probabilities $P(S_j|\theta_j)$ the bound has a lower limit and therefore cannot follow closely the true marginal probability for very improbable events.

To simplify the minimization with respect to $\xi$ we can work on a log scale and make use of the following Legendre transformation:

$$\log x = \min_{\lambda}\{\lambda x - \log\lambda - 1\} \quad (13)$$

As a result we get

$$\log P(\{S_i\}_{i\in L_1}|\theta) \leq -\sum_{i\in L_1} H(\xi_i) + \sum_{j\in L_2}\lambda_j\left(P(S_j=1|\theta_j)e^{\sum_{i\in L_1}\xi_i(2S_i-1)\theta_{ij}} + P(S_j=0|\theta_j)\right) + \sum_{j\in L_2}[-\log\lambda_j - 1] \quad (14)$$

where we have ceased to indicate explicitly that the bound will be minimized over the adjustable parameters. This new form of the bound has the advantage that the minimization with respect to each parameter ($\xi$ or $\lambda$) is reduced to convex optimization[1] and can be done by any standard method (e.g. Newton-Raphson). The convexity property is important in guaranteeing a unique and accessible minimum for any of the variational parameters at each step of the sequential (iterative) optimization. Note that the accuracy of the bound is not compromised by the additional Legendre transformation. Its effect is merely to simplify the expressions for optimization.

[1]The convexity with respect to each $\xi$ follows from the convexity of $e^x$ and the positivity of the multiplying coefficients $\lambda$.

## 2.2 GENERIC LOWER BOUND FOR SIGMOID NETWORK

Methods for finding lower bounds on marginal likelihoods were first presented by Dayan, et al. (1995) and Hinton, et al. (1995) in the context of a layered network known as the "Helmholtz machine". Saul, et al. (1996) subsequently provided a rigorous calculation of lower bounds (by appeal to mean field theory) in the case of generic sigmoid networks. Unlike the method for obtaining upper bounds presented in the previous section, the lower bound methodology poses no constraints on the network structure. We briefly introduce the idea here (for more details see Saul, et al.).

Let us denote the marginal set of variables by $\{S_i\}_{i\in L}$. A lower bound on the (log) marginal probability can be found directly via Jensen's inequality:

$$\log P(\{S_i\}_{i\in L}|\theta) = = \log\sum_{\{S\}_{i\notin L}} P(S_1,\ldots,S_n|\theta)$$

$$= \log\sum_{\{S_i\}_{i\notin L}} Q(\{S\})\frac{P(S_1,\ldots,S_n|\theta)}{Q(\{S\})}$$

$$\geq \sum_{\{S\}_{i\notin L}} Q(\{S\})\log\frac{P(S_1,\ldots,S_n|\theta)}{Q(\{S\})} \quad (15)$$

which holds for any distribution $Q$ over $\{S_i\}_{i\notin L}$. The bound becomes exact if $Q(\{S\})$ can represent the true posterior distribution $P(\{S\}\mid\{S_i\}_{i\in L},\theta)$. For other choices of $Q$ the accuracy of the bound is characterized by the Kullback-Leibler distance between $Q$ and the posterior. As we are assuming that computing the likelihood exactly is intractable the idea is to find a distribution $Q$ that can be computed efficiently. The simplest of such distributions is the completely factorized ("mean field") distribution:

$$Q(\{S\}) = \prod_j \mu_j^{S_j}(1-\mu_j)^{1-S_j} \quad (16)$$

Inserting this distribution into the lower bound (Eq. (15)) we can, in principle, carry out the summation[2] and get an expression for the lower bound. Consequently, the adjustable variational parameters $\mu_j$ can be modified to make the bound tighter.

For later utility we rewrite the lower bound in eq. (15) as[3]:

$$\log P(\{S_i\}_{i\in L}|\theta) \geq E_Q\{\ \log P(S_1,\ldots,S_n|\theta)\ \} + H_Q \quad (17)$$

$$= \sum_i E_Q\{\ \log P(S_i|\mathrm{pa}[i],\theta)\ \} + H_Q \quad (18)$$

where $H_Q$ is the entropy of the $Q$ distribution and $E_Q\{\cdot\}$ is the expectation with respect to $Q$. At this point we have proceeded as far as possible for generic architectures; further development of the bound is dependent on the type of the network – whether sigmoid, noisy-OR, or other[4].

[2]The summation even in case of simple factorized distributions can be non-trivial to perform; see Saul, et al.

[3]$\sum Q\log P/Q = \sum Q\log P + (-\sum Q\log Q)$

### 2.3 NUMERICAL EXPERIMENTS FOR SIGMOID NETWORK

In testing the accuracy of the developed bounds we used $8 \rightarrow 8$ networks (complete bipartite graphs as in Figure 1 with 8 nodes in each level), where the network size was chosen to be small enough to allow exact computation of the marginal probabilities for purposes of comparison. The method of testing was as follows. The parameters for the $8 \rightarrow 8$ networks were drawn from a Gaussian prior distribution and a sample from the resulting joint distribution of the variables was generated. The variables in the "receiving" layer of the bipartite graph were set according to the sample. The true marginal probability as well as the upper and lower bounds were computed for this setting. The resulting bounds were assessed by employing the relative error in log-likelihood, i.e. $(\log P_{\text{Bound}}/\log P - 1)$, as a measure of accuracy.

More precisely, the prior distribution over the parameters was taken to be

$$P(\theta) = \prod_i \prod_{j \in \text{pa}[i]} \frac{1}{\sqrt{2\pi\sigma^2}} e^{-\frac{1}{2\sigma^2}\theta_{ij}^2} \quad (19)$$

where the overall variance $\sigma^2$ allows us to vary the degree to which the resulting parameters make the two layers of the network dependent. For small values of $\sigma^2$ the layers are almost independent whereas larger values make them strongly interdependent. To make the situation worse for the bounds[5] we enhanced the coupling of the layers by setting $P(S_j|\theta_j) = 1/2$ for the variables not in the desired marginal set, i.e., making them maximally variable.

In order to make the accuracy of the bounds commensurate with those for the noisy-OR networks reported below, we summarize the results via a measure of interlayer dependence. This dependence was estimated by

$$\sigma_{std} = \max_{i \in L_1} \sqrt{\text{Var}\{P(S_i|\text{pa}[i])\}} \quad (20)$$

i.e., the maximum variability of the conditional likelihoods. Here $S_i$ was fixed in the $P(S_i|\text{pa}[i])$ functional according to the initial sample and the variance was computed with respect to the joint distribution[6].

Figure 2 illustrates the accuracy of the bounds as measured by the relative log-likelihood as a function of $\sigma_{std}$[7]. In terms of probabilities, a relative error of $\epsilon$ translates into a $P^{1+\epsilon}$ approximation of the true likelihood $P$. Note that the relative error is always positive for the upper bound and negative for the lower bound, as guaranteed by the theory. The figure indicates that the bounds are accurate enough to be useful. In addition, we see that the the upper bound deteriorates faster with increasingly coupled layers.

Let us now briefly consider the scaling properties of the bounds as the network size increases. We note first that the evaluation time for the bounds increases approximately linearly with the number of parameters $\theta$ in these two-level networks[8]. The accuracy of the bounds, on the other hand, needs experimental illustration.

In large networks it is not feasible to compute $\sigma_{std}$ nor the true marginal likelihood. We may, however, calculate the relative error between the upper and lower bounds. To maintain approximately same level of $\sigma_{std}$ across different network sizes we plotted the errors against $\sigma\sqrt{n}$ (for fully connected $n$ by $n$ two-level networks), where $\sigma$ is the overall standard deviation in the prior distribution. Figure 3 shows that the relative errors are invariant to the network size in this scaling.

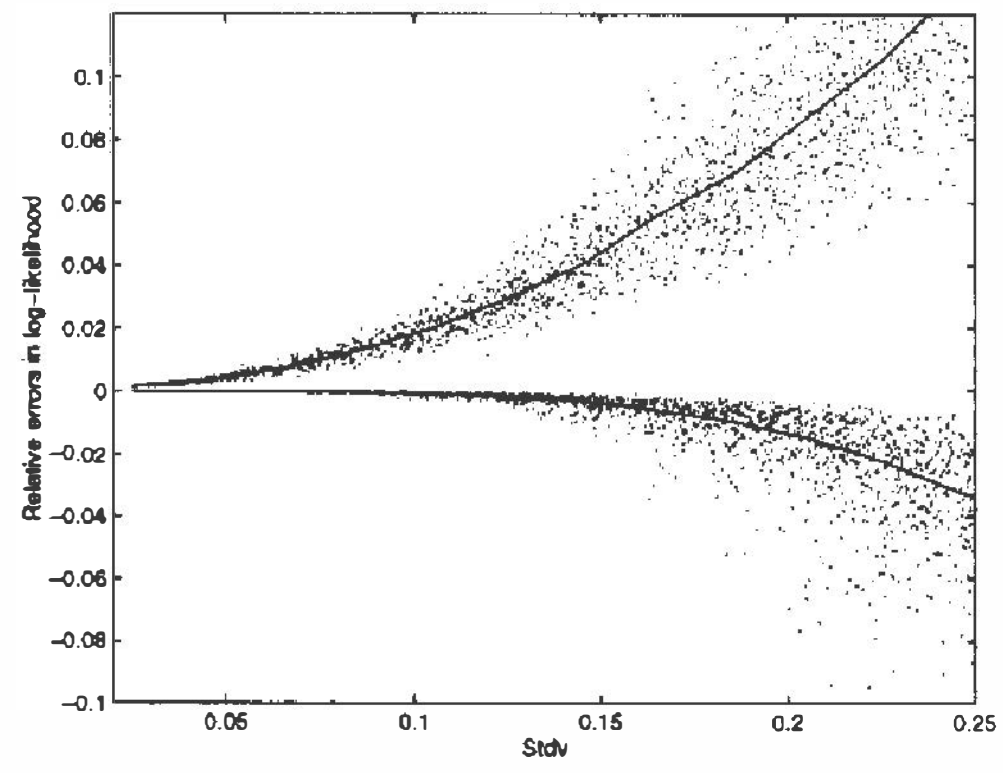

Figure 2: Sigmoid networks. Accuracy of the bounds for 8 by 8 two-level networks. The solid lines are the median relative errors in log-likelihood as a function of $\sigma_{std}$. The upper and lower curves correspond to the upper and lower bounds respectively.

## 3 NOISY-OR NETWORK

Noisy-OR networks – like sigmoid networks – can be represented by DAGs and are written as a product

[4] For a derivation of lower bounds for networks with cumulative Gaussians replacing the sigmoid function see Jaakkola et al. (1996).

[5] Both the upper and lower bounds are exact in the limit of lightly coupled layers.

[6] Note that $P(S_i|\text{pa}[i])$ with $S_i$ fixed is just some function of the variables in the network whose variance can be computed.

[7] Note that the maximum value for $\sigma_{std}$ is $1/2$.

[8] The amount of computation needed for sequentially optimizing each variational parameter once scales linearly with the number of network parameters. Only a few such iterations are needed.

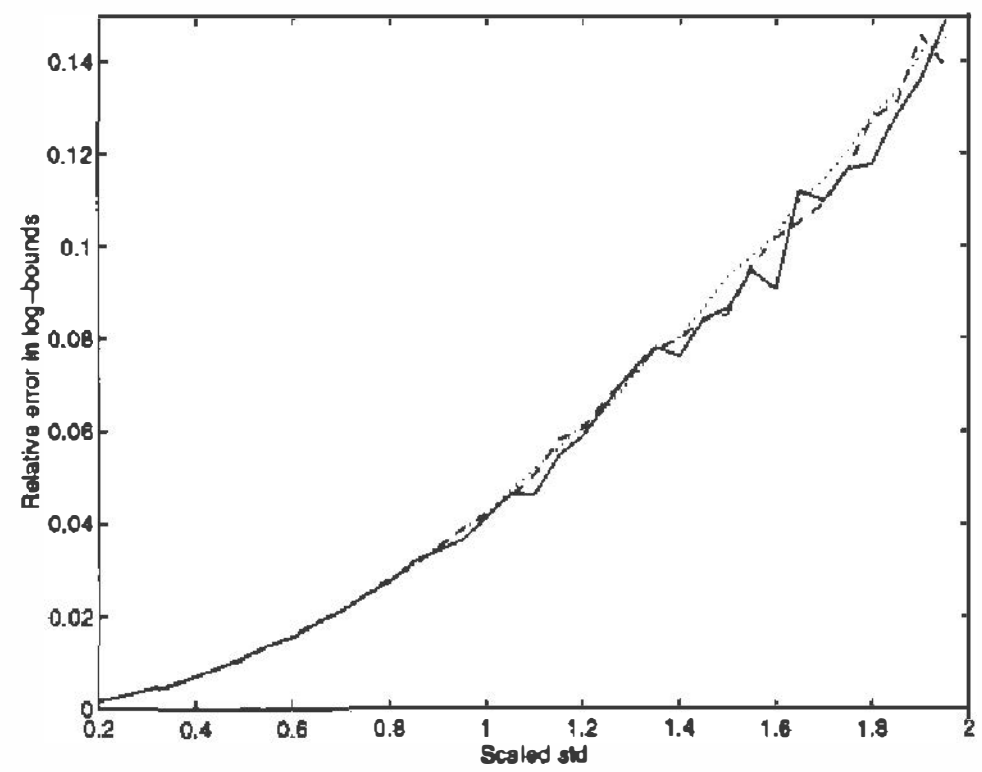

Figure 3: Sigmoid networks. Median relative errors between the upper and lower bounds (log scale) as a function of $\sigma\sqrt{n}$ for $n$ by $n$ two-level networks. Solid line: $n = 8$; dashed line: $n = 32$; dotted line: $n = 128$.

form for the joint distribution:

$$P(S_1,\ldots,S_n|\theta) = \prod_i P(S_i|\mathrm{pa}[i],\theta) \qquad (21)$$

Unlike sigmoid networks, however, the conditional probabilities for a noisy-OR network are defined as

$$\begin{aligned} P(S_i|\mathrm{pa}[i],\theta) &= \left(1 - \prod_{j\in \mathrm{pa}[i]} (1-q_{ij})^{S_j}\right)^{S_i} \\ &\times \left(\prod_{j\in \mathrm{pa}[i]} (1-q_{ij})^{S_j}\right)^{1-S_i} \qquad (22) \end{aligned}$$

where, for example, the parameter $q_{ij}$ corresponds to the probability that the $j^{th}$ parent of $i$ alone can turn $S_i$ on. A constant "leak" or "bias" can be included by introducing a dummy (parent) variable whose value is always fixed to one.

In the following two sections we develop methods for computing upper and lower bounds on marginal probabilities for noisy-OR networks. Similarly to the case of sigmoid networks the upper bound is applicable to a restricted class of networks while the lower bound remains generic. For clarity of the forthcoming derivations we introduce the notation:

$$\begin{aligned} P(S_i=0|\mathrm{pa}[i],\theta) &= \prod_{j\in \mathrm{pa}[i]} (1-q_{ij})^{S_j} \\ &= e^{-\sum_{j\in \mathrm{pa}[i]}\theta_{ij}S_j} \qquad (23) \end{aligned}$$

with $\theta_{ij} = -\log(1-q_{ij}) \geq 0$.

## 3.1 UPPER BOUND FOR NOISY-OR NETWORK

The motivation and, in broad outline, the upper bound derivation itself can be carried over from the sigmoid setting to the noisy-OR case.

Consider a two-level or bipartite network with $\{S_i\}_{i\in L_1}$ and $\{S_i\}_{i\in L_2}$ (where $L_2 \rightarrow L_1$) denoting the two sets of variables. As before we adopt a simplifying notation in which the desired marginal probability exclusively consists of all the variables in the layer $L_1$. To compute such marginal we need to sum the noisy-OR joint distribution,

$$\begin{aligned} &P(S_1,\ldots,S_n|\theta) = \\ &= \prod_{i\in L_1} (1-e^{-\sum_j \theta_{ij}S_j})^{S_i} e^{-(1-S_i)\sum_j \theta_{ij}S_j} \\ &\quad \times \prod_{j\in L_2} P(S_j|\theta_j) \qquad (24) \end{aligned}$$

over the variables in $L_2$. We note that in case of fully connected bipartite networks the complexity of performing this calculation increases exponentially with the number of variables in $L_1$ that are set to one (D'Ambrosio, 1994); importantly, and unlike in the sigmoid case, the complexity does not vary exponentially with the number of marginalized variables. Nevertheless, we focus on the case where the exact method of obtaining the marginal probability is infeasible.

To find an upper bound in the noisy-OR setting we use the following variational transformation (for a derivation and discussion see appendix B)

$$1 - e^{-x} = \min_{\xi\geq 0} e^{\xi x - F(\xi)} \qquad (25)$$

where $F(\xi) = -\xi\log\xi + (\xi+1)\log(\xi+1)$. By inserting this transformation into the joint distribution we obtain:

$$\begin{aligned} &P(S_1,\ldots,S_n|\theta) = \\ &= \prod_{i\in L_1} \min_{\xi_i}\left\{ e^{S_i[\sum_j \theta_{ij}S_j - F(\xi_i)]}\right\} e^{-(1-S_i)\sum_j \theta_{ij}S_j} \\ &\quad \times \prod_{j\in L_2} P(S_j|\theta_j) \qquad (26) \\ &= \min_{\xi}\left\{ e^{-\sum_{i\in L_1} S_i F(\xi_i)} \times \right. \\ &\quad \left. \prod_{j\in L_2} \left[ e^{\sum_{i\in L_1}(S_i\xi_i + S_i - 1)\theta_{ij}} \right]^{S_j} P(S_j|\theta_j) \right\} \qquad (27) \\ &\stackrel{def}{=} \min_{\xi}\{\, \tilde{P}(S_1,\ldots,S_n|\theta,\xi)\,\} \qquad (28) \end{aligned}$$

where we have regrouped terms by rewriting the product over $i \in L_1$ as a sum in the exponent and collecting the terms depending on the variables $\{S_j\}_{j\in L_2}$. We can see that the implicitly defined (and unnormalized) $\tilde{P}(S_1,\ldots,S_n|\theta,\xi)$ factorizes over $S_j$. As in the sigmoid case, this factorial property allows us to find a closed form upper bound on the marginal:

$$\begin{aligned} &P(\{S_i\}_{i\in L_1}|\theta) = \\ &= \sum_{\{S_j\}_{j\in L_2}} P(S_1,\ldots,S_n|\theta) \end{aligned}$$

$$= \sum_{\{S_j\}_{j\in L_2}} \min_{\xi} \tilde{P}(S_1,\ldots,S_n|\theta,\xi) \quad (29)$$

$$\leq \min_{\xi} \sum_{\{S_j\}_{j\in L_2}} \tilde{P}(S_1,\ldots,S_n|\theta,\xi) \quad (30)$$

where the last summation can now be performed exactly to yield:

$$P(\{S_i\}_{i\in L_1}|\theta) \leq \min_{\xi} \left\{ e^{-\sum_{i\in L_1} S_i F(\xi_i)} \times \prod_{j\in L_2} \left( P(S_j=1|\theta_j) e^{\sum_{i\in L_1}(S_i\xi_i+S_i-1)\theta_{ij}} + P(S_j=0|\theta_j) \right) \right\} \quad (31)$$

This bound (i) always stays below (or equal to) one as it is less than or equal to one whenever all $\xi$ are set to zero, and (ii) is exact when all $S_i$ in $L_1$ are zero or in the limit of vanishing parameters $\theta_{ij}$.

Similarly to the sigmoid case we may simplify the minimization process by considering $\log P(\{S_i\}_{i\in L_1}|\theta)$ and introducing a Legendre transformation for $\log(\cdot)$. This yields:

$$\log P(\{S_i\}_{i\in L_1}|\theta) \leq \sum_{i\in L_1} S_i F(\xi_i) + \sum_{j\in L_2} \lambda_j \left( P(S_j=1|\theta_j) e^{\sum_{i\in L_1}(S_i\xi_i+S_i-1)\theta_{ij}} + P(S_j=0|\theta_j) \right) + \sum_{j\in L_2} [-\log\lambda_j - 1] \quad (32)$$

where we have dropped the explicit reference to minimization. The gain again is the convexity of the bound with respect to any of the $\xi$ or $\lambda$ variables.

## 3.2 GENERIC LOWER BOUND FOR NOISY-OR NETWORK

The earlier work on lower bounds by Saul, et al. was restricted to sigmoid networks; we extend that work here by deriving a lower bound for generic noisy-OR networks. We refer to section 2.2 for the framework and commence from the noisy-OR counterpart of eq. (18). Thus,

$$\log P(\{S_i\}_{i\in L}|\theta) \geq \sum_i E_Q\{ \log P(S_i|\text{pa}[i],\theta) \} + H_Q \quad (33)$$

$$= \sum_i E_Q\{ S_i \log(1 - e^{-\sum_j \theta_{ij} S_j}) \} + \sum_i E_Q\{ -(1-S_i)\sum_j \theta_{ij} S_j \} + H_Q \quad (34)$$

which is obtained by writing explicitly the form of the conditional probabilities for noisy-OR networks. While the second expectation in eq. (34) simply corresponds to replacing the binary variables $S_i$ with their means $\mu_i$ (since $Q$ is factorized), the first expectation lacks a closed form expression. To compute this expectation efficiently we make use of the following expansion:

$$1 - e^{-x} = \prod_{k=0}^{\infty} g(2^k x) \quad (35)$$

where $g(\cdot)$ is the sigmoid function (see appendix C). This expansion converges exponentially fast and thus only a few terms need to be included in the product for good accuracy. By carrying out this expansion in the bound above and explicitly using the form of the sigmoid function we get

$$\log P(\{S_i\}_{i\in L}|\theta) \geq \sum_i \sum_k E_Q\{ -S_i \log(1 + e^{-2^k \sum_j \theta_{ij} S_j}) \} - \sum_i (1-\mu_i) \sum_j \theta_{ij}\mu_j + H_Q \quad (36)$$

Now, as the parameters $\theta_{ij}$ are non-negative,

$$e^{-2^k \sum_j \theta_{ij} S_j} \in [0,1]$$

and we may use the smooth convexity properties of $-\log(1+x)$ (for $x \in [0,1]$) to bring the expectations in eq. (36) inside the log. This results in

$$\log P(\{S_i\}_{i\in L}|\theta) \geq \sum_{ik} -\mu_i \log \left[ 1 + \prod_j (\mu_j e^{-2^k \theta_{ij}} + 1 - \mu_j) \right] - \sum_i (1-\mu_i) \sum_j \theta_{ij}\mu_j + H_Q \quad (37)$$

A more sophisticated and accurate way of computing the expectations in eq. (36) is discussed in appendix D.

## 3.3 NUMERICAL EXPERIMENTS FOR NOISY-OR NETWORK

The method of testing used here was, for the most part, identical to the one presented earlier for sigmoid networks (section 2.3). The only difference was that the prior distribution over the parameters defining the conditional probabilities was chosen to be a Dirichlet instead of a Gaussian:

$$q_{ij} \sim \phi(1-q_{ij})^{\phi-1} \quad (38)$$

(recall that $P(S_i = 0|\text{pa}[i],\theta) = \prod_{j\in\text{pa}[i]}(1-q_{ij})^{S_j}$). For large $\phi$, $q$ stays small (or $1 - q \approx 1$) and the layers of the bipartite network are only weakly connected; smaller values of $\phi$, on the other hand, make the layers strongly dependent. We thus used $\phi$ to vary (on average) the interdependence between the two layers. To facilitate comparisons with the bounds derived for sigmoid networks we used $\sigma_{std}$ (see eq. (20)) as a measure of dependence between the layers.

Figure 4 illustrates the accuracy of the computed bounds as a function of $\sigma_{std}$[9]. The samples with zero relative error are from the upper/lower bounds in cases where all the variables in the desired marginal are zero since the bounds become exact whenever this happens. The lower bound is slightly worse than the one for sigmoid networks most likely due to the symmetry and smoother nature of the sigmoid function. As with the sigmoid networks the upper bound becomes less accurate more quickly.

We now turn to the effects of increasing the network size. Analogously to the sigmoid networks the evaluation times for the bounds vary approximately linearly with the number of parameters in these two-level networks, albeit with slightly larger coefficients (for the lower bound). As for the accuracy of the bounds, Figure 5 shows the relative errors[10] between the bounds across different network sizes. The errors are plotted against $\sqrt{n}/\phi$ for $n$ by $n$ two-level networks, where $\phi$ defines the Dirichlet prior distribution for the parameters. In the chosen scale the network size has little effect on the errors[11].

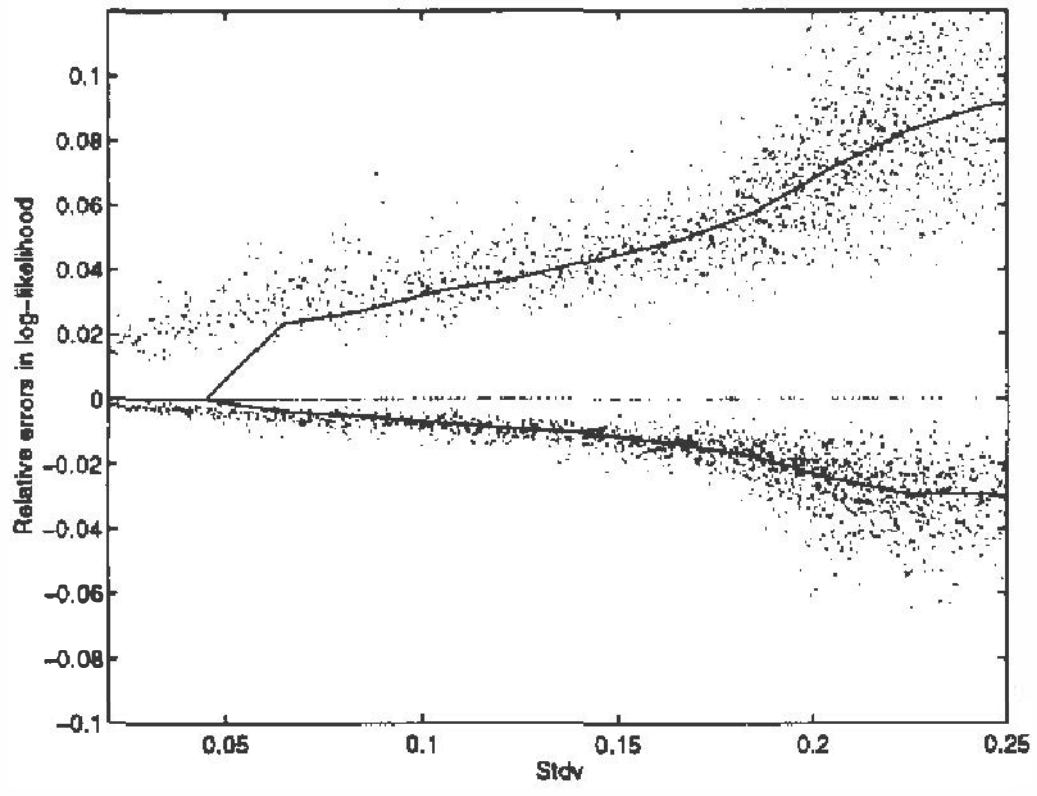

Figure 4: Noisy-OR network. Accuracy of the bounds for 8 by 8 two-level networks. The solid lines are the median relative errors in log-likelihood as a function of $\sigma_{std}$. The upper and lower curves correspond to the upper and lower bounds respectively.

## 4 DISCUSSION AND FUTURE WORK

Applying probabilistic methods to real world inference problems can lead to the emergence of cliques that are prohibitively large for exact algorithms (for example, in medical diagnosis). We focused on dealing with such large (sub)structures in the context of sigmoid belief networks and noisy-OR networks. For these networks we developed techniques for computing upper and lower bounds on marginal probabilities. The bounds serve as an alternative to sampling methods in the presence of intractable structures. They can define interval bounds for the marginals and can be used to improve the accuracy of decision making in intractable networks.

Toward extending the work presented in this paper we note that both the upper and lower bounds can be improved by considering a mixture partitioning (Jaakkola & Jordan, 1996) of the space of marginalized variables instead of using a completely factorized approximation. Furthermore, the restriction of the upper bounds for two-level networks can be overcome, for example, by interlacing them with sampling techniques, although other extensions may be possible as well. Following Saul & Jordan (1996) we may also merge the obtained bounds with exact methods whenever they are feasible.

[9] The slight unevenness of the samples are due to the non-linear relationship between the Dirichlet parameter $\phi$ and $\sigma_{std}$.

[10] The errors are for the worst case marginal, i.e., for $P(\{S_i = 1\}_{i \in L_1})$.

[11] The 8 by 8 network is too small to be in the desired asymptotic regime.

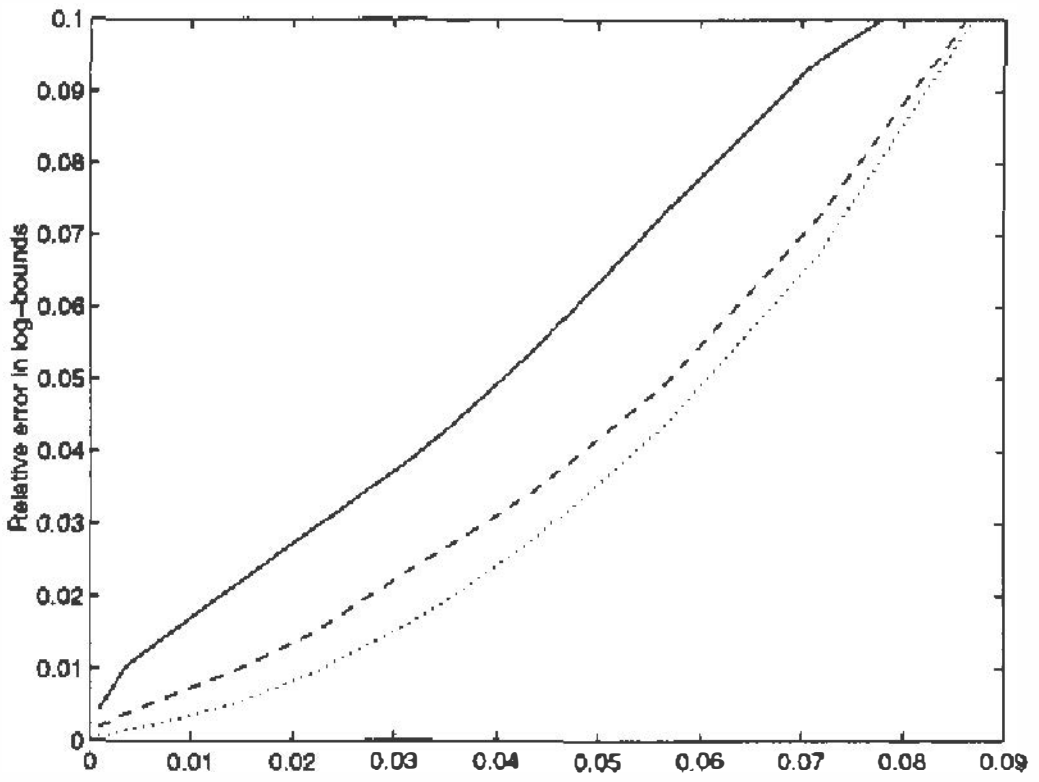

Figure 5: Noisy-OR network. Median relative errors between the upper and lower bounds (in log scale) as a function of $\sqrt{n}/\phi$ for $n$ by $n$ two-level networks. Solid line: $n = 8$; dashed line: $n = 32$; dotted line: $n = 128$.

## Acknowledgments

The authors wish to thank L. K. Saul and the anonymous reviewers for helpful comments and suggestions.

# A SIGMOID TRANSFORMATION

Here we derive and discuss the following transformation:

$$g(x) = \frac{1}{1+e^{-x}} = \min_{\xi\in[0,1]} e^{\xi x - H(\xi)}$$

Although a proof by hindsight would be shorter than a direct derivation we present the derivation for it is more informative. To this end, let us switch to log scale and consider

$$\begin{aligned}
-\log(1+e^{-x}) &= -\log \sum_{m\in\{0,1\}} e^{-mx} \\
&= -\log \sum_{m\in\{0,1\}} \xi^m (1-\xi)^{1-m} \frac{e^{-mx}}{\xi^m(1-\xi)^{1-m}} \\
&= -\log E\{\frac{e^{-mx}}{\xi^m(1-\xi)^{1-m}}\} \\
&\leq E\{-\log \frac{e^{-mx}}{\xi^m(1-\xi)^{1-m}}\} \\
&= \xi x + \xi \log \xi + (1-\xi)\log(1-\xi) \\
&= \xi x - H(\xi)
\end{aligned}$$

which follows from interpreting $\xi^m(1-\xi)^{1-m}$ as a probability mass for $m$ and from an application of Jensen's inequality. By actually performing the minimization over $\xi$ gives $\xi^* = g(-x)$ and leads to an equality instead of a bound. The geometry of the bound when $\xi$ is kept fixed for all $x$ is illustrated in figure 6. The value of $x$ for which the chosen $\xi$ is optimal is the point where the bound is exact.

We finally note that the above transformation can be understood as a type of Legendre transformation.

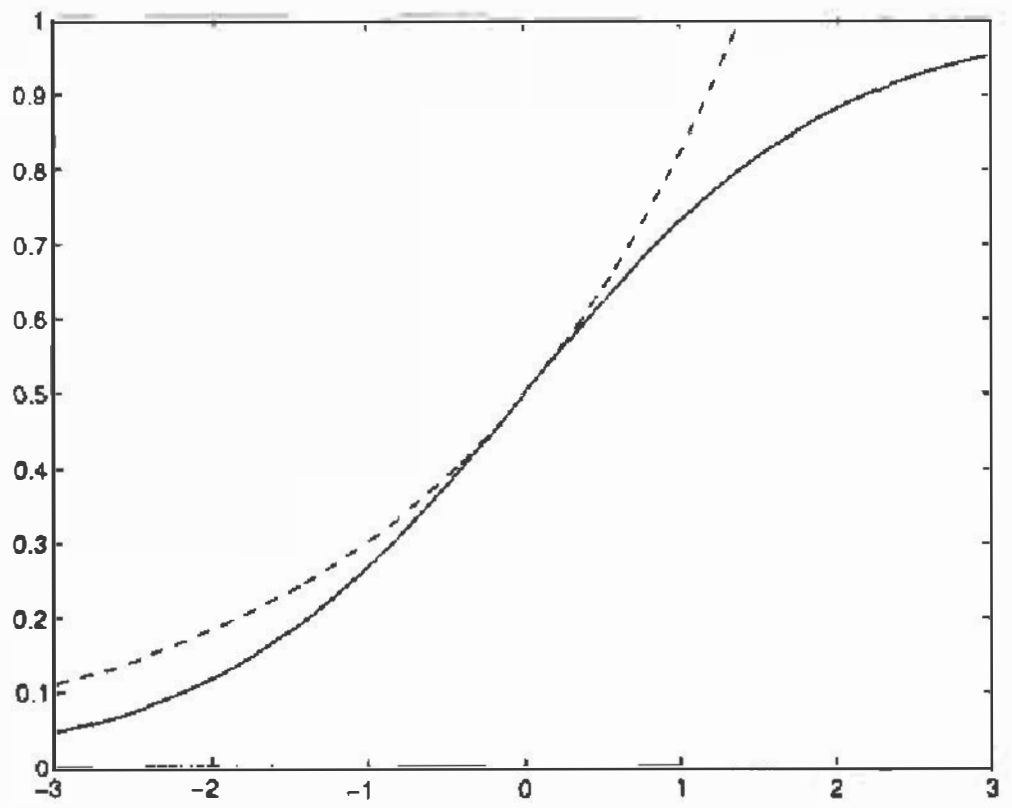

Figure 6: Geometry of the sigmoid transformation. The dashed curve plots $\exp\{\xi x - H(\xi)\}$ as a function of $x$ for a fixed $\xi$ (=0.5).

# B NOISY-OR TRANSFORMATION

Here we provide a derivation for the transformation

$$1 - e^{-x} = \min_{\xi\geq 1} e^{\xi x - F(\xi)} \tag{39}$$

presented in the text. Switching to log scale we find

$$\begin{aligned}
\log(1-e^{-x}) &= -\log\frac{1}{1-e^{-x}} = -\log \sum_{k=0}^{\infty} e^{-kx} \\
&= -\log \sum_{k=0}^{\infty} (1-q)q^k \frac{e^{-kx}}{(1-q)q^k} \\
&= -\log E\{\frac{e^{-kx}}{(1-q)q^k}\} \\
&\leq E\{-\log \frac{e^{-kx}}{(1-q)q^k}\}
\end{aligned}$$

$$= \sum_{k=0}^{\infty}(1-q)q^k kx + \sum_{k=0}^{\infty}(1-q)q^k[\log(1-q)+k\log q]$$
$$= \frac{q}{1-q}x + \log(1-q) + \frac{q}{1-q}\log q$$

where we have interpreted $(1-q)q^k$ as a probability distribution for $k$ and used Jensen's inequality. Minimizing the above bound with respect to $q$ gives $q^* = e^{-x}$ and the bound becomes exact. The original transformation follows by setting $\xi = q/(1-q)$. If the value of $\xi$ is kept constant, the transformation yields a bound, the geometry of which is shown in figure 7. The point where the bound touches the $1-e^{-x}$ curve defines $x$ for which the constant $\xi$ is optimal.

As in the sigmoid case the resulting transformation can be seen as a type of Legendre transformation.

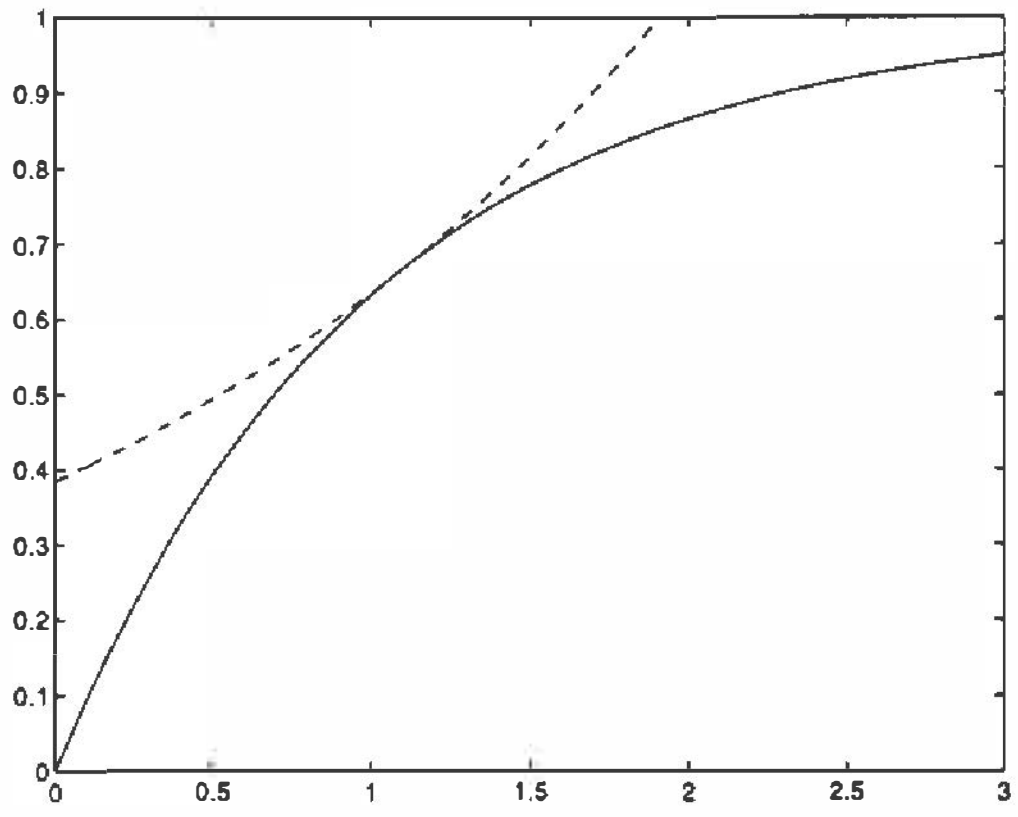

Figure 7: Geometry of the noisy-OR transformation. The dashed curve gives $\exp\{\xi x - F(\xi)\}$ as a function of $x$ when $\xi$ is fixed at 0.5.

# C NOISY-OR EXPANSION

The noisy-OR expansion

$$1-e^{-x} = \prod_{k=0}^{\infty} g(2^k x) \tag{40}$$

follows simply from

$$\begin{aligned} 1-e^{-x} &= \frac{(1+e^{-x})(1-e^{-x})}{1+e^{-x}} \\ &= g(x)(1-e^{-2x}) \\ &= g(x)\frac{(1+e^{-2x})(1-e^{-2x})}{1+e^{-2x}} \\ &= g(x)g(2x)(1-e^{-4x}) \end{aligned} \tag{41}$$

and induction. For $x > 0$ the accuracy of the expansion is governed by $1-e^{-2^k x}$ which goes to one exponentially fast. Also since $g(2^k 0) = 1/2$, the expansion becomes $(\frac{1}{2})^N$ at $x = 0$, where $N$ is the number of terms included. As this approaches $1-e^{-0} = 0$ exponentially fast, we conclude that the rapid convergence is uniform. Figure 8 illustrates the accuracy of the expansion for small $N$.

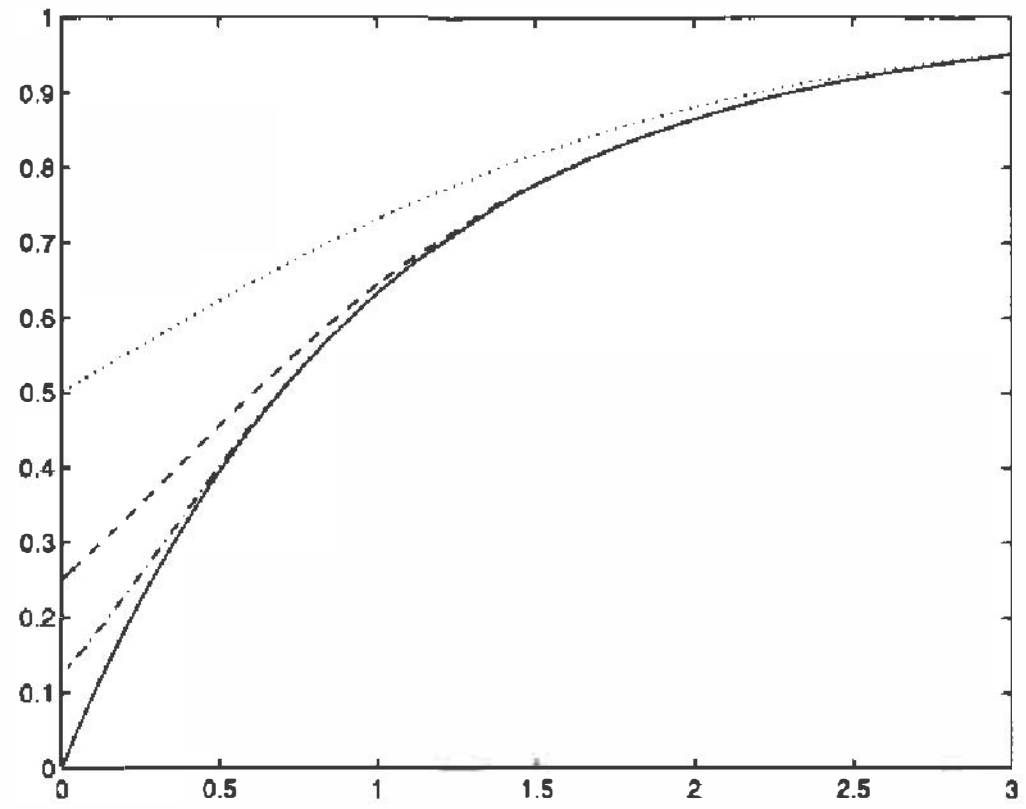

Figure 8: Accuracy of the noisy-OR expansion. Dotted line: $N = 1$; dashed line: $N = 2$; dotdashed: $N = 3$. $N$ is the number of terms included in the expansion.

# D QUADRATIC BOUND

For $X \in [0,1]$ we can bound $-\log(1+X)$ by a quadratic expression:

$$-\log(1+X) \geq a(X-x)^2 + b(X-x) + c \tag{42}$$

where $c = -\log(1+x)$, $b = -1/(1+x)$, and $a = -[(1-x)b + c + \log 2]/(1-x)^2$. The coefficents can be derived by requiring that the quadratic expression and its derivative are exact at $X = x$, and by choosing the largest possible $a$ such that the expression remains a bound. The resulting approximation is good for all $x \in [0,1]$ and can be optimized by setting $x = E\{X\}$.

Let us now use this quadratic bound in eq. (36) to better approximate the expectations. To simplify the ensuing formulas we use the notation

$$E_Q\{\, e^{-2^k \sum_j \theta_{ij} S_j}\,\} = \;= \prod_j \left(\mu_j e^{-2^k \theta_{ij}} + 1 - \mu_j\right) = X_i^{(k)} \tag{43}$$

With these we straightforwardly find

$$\begin{aligned} &\log P(\{S_i\}_{i\in L}|\theta) \\ &\geq \sum_{ik} \mu_i a_{ik}\left[X_i^{(k+1)} - 2X_i^{(k)}x_i^{(k)} + (x_i^{(k)})^2\right] \\ &\quad + \sum_{ik}\mu_i\left[b_{ik}(X_i^{(k)} - x_i^{(k)}) + c_{ik}\right] \\ &\quad - \sum_i (1-\mu_i)\sum_j \theta_{ij}\mu_j + H_Q \end{aligned} \tag{44}$$

which is optimized with respect to $x_i^{(k)}$ simply by setting $x_i^{(k)} = X_i^{(k)}$. The simpler bound in eq. (37) corresponds to ignoring the quadratic correction, i.e., using $a_{ik} = 0$ above.